\def\year{2022}\relax
\pdfoutput=1
\documentclass[letterpaper]{article} 
\usepackage{aaai22}  
\usepackage{times}  
\usepackage{helvet}  
\usepackage{courier}  
\usepackage[hyphens]{url}  
\usepackage{graphicx} 
\usepackage{natbib}  
\usepackage{caption} 
\DeclareCaptionStyle{ruled}{labelfont=normalfont,labelsep=colon,strut=off} 
\graphicspath{ {Figures/}}
\usepackage{algorithm}
\usepackage{algorithmic}

\usepackage{comment}
\usepackage{lipsum}
\usepackage{booktabs}
\usepackage{threeparttable}
\usepackage{newfloat}
\usepackage{listings}
\floatstyle{ruled}
\newfloat{listing}{tb}{lst}{}
\floatname{listing}{Listing}
\title{Leveraging Centric Data Federated Learning Using Blockchain For Integrity Assurance}
\author{
   Riadh Ben Chaabene,\textsuperscript{\rm 1}
    Darine Amayed, \textsuperscript{\rm 1}
    Mohamed Cheriet \textsuperscript{\rm 1}
}
\affiliations{
    \textsuperscript{\rm 1}École de technologie supérieure ÉTS\\
    \textsuperscript{\rm 1}Systems Engineering Department (Synchromedia lab)\\

    21100 Notre-Dame St W\\
    Montreal, Quebec H3C 1K3\\
    riadh.ben-chaabene.1@ens.etsmtl.ca,
    darine.ameyed.1@ens.etsmtl.ca,
    mohamed.cheriet@etsmtl.ca

}

\begin{document}

\maketitle

\begin{abstract}
Machine learning abilities have become a vital component for various solutions across industries, applications, and sectors. Many organizations seek to leverage AI-based solutions across their business services to unlock better efficiency and increase productivity. Problems, however, can arise if there is a lack of quality data for AI-model training, scalability, and maintenance.  
We propose a data-centric federated learning architecture leveraged by a public blockchain and smart contracts to overcome this significant issue. Our proposed solution provides a virtual public marketplace where developers, data scientists, and AI-engineer can publish their models and collaboratively create and access quality data for training. We enhance data quality and integrity through an incentive mechanism that rewards contributors for data contribution and verification. Those combined with the proposed framework helped increase with only one user simulation the training dataset with an average of 100 input daily and the model accuracy by approximately 4\%.

\end{abstract}
\section{Introduction}
The machine learning market is witnessing tremendous growth. In 2020, the market saw an increase of 38.01\%, leading to a compound annual growth rate growth of 39\%  and an investment intercepted at \$11.16 Billion between the year 2020 and 2024 \cite{Technavio}. This is due to the fragmentation of the market with large companies investing and contributing to the artificial intelligence field, such as Alphabet Inc, Amazon Inc, and many others. Moreover, in the last few years, machine learning research has augmented tremendously due to its contribution to numerous areas, making them more robust and systematic. Society and companies have benefited from this progress, leading to multiple investments to encourage this growth. Researchers and developers have been dedicated to providing more studies and contributions to this technology. Despite all of this, machine learning is still facing significant challenges; two of them consist of data accessibility and trustability. 
Data \cite{biderman_pitfalls_2021} is the key that is driving the majority of machine learning models. Acutely, the data is facing the following issues: high centralization, complex accessibility, insufficient security, and insufficient quality.
Companies and research centers are acting with the problem of acquiring the correct data for their research\cite{biderman_pitfalls_2021}, since most of the data tends to be centralized and inaccessible, leading to either unavailability or high cost of acquirement. Also, data can easily be tampered with, which causes a lack of data quality. Even if the data is available, noisy information could affect the whole model. Results of a model could variety greatly from one dataset to another, and since the data commonly being used is variable, there is no guarantee that the data is equitable.\\ 
Additionally, creating the model relies on data collection, which consists of 60\% of the overall machine learning work and that 25\% of the time is allocated for data cleaning and data labeling\cite{roh_survey_2019}. Despite the importance of data, 92\% of it is owned and stored in the western world \cite{oliverwyman}, where most of the big companies lie. 
On the other hand, machine learning model training is dependent on human interaction, causing the learning process to start or end in an insignificant time leading to the problem of automation.\\
It is therefore mandatory to provide alternative solutions to these problems. They will enhance the progress of machine learning, leading to the creation of a more efficient and sophisticated model to improve the development of this technology and our lives.
All of this leads to the categorization the problem into three areas:
\begin{itemize}
    \item Model training issues: that consist of the insufficient data and resources for model training, leading to less accuracy.
    \item Data availability issues: that corresponds to the lack of data available for people to train machine learning models.
    \item Data Integrity issues: that refer to the insurance of the available data quality.
\end{itemize}

In this paper, we propose a framework for publishing and contributing to the collaborative machine learning process. The idea is to present a public marketplace with free access to share a model on the one end and contribute to the continuous training of this model on the other. We encourage participants to share and verify the data to provide the needed data quality for the training to avoid manipulation. Also, we are looking for model diversity where this framework will function with any machine learning model type. To reach our objective, we will be using a collaborative learning approach such as federated learning. Furthermore, we will be using blockchain technology \cite{nakamoto_bitcoin_nodate} as an infrastructure for our framework and the InterPlanetary File System as a storage mechanism. Also, a monetary incentive mechanism will be developed to reward contributors and withhold malicious users.
\section{Background}
Federated Learning \cite{yang_federated_2019} conducts machine learning in a decentralized approach. It aims to provide the model to the data rather than the traditional way, where we create a model then provide the data. The data is distributed along with different users in their edge mobiles. Each provides his data to train the model, leading to multiple unique training datasets rather than one. Only the metadata is returned to the centralized model or the global model using encrypted communication. At the same time, this provides knowledge performance for the model and enhances data privacy since all the calculations happen inside the owner's device without any exchange of their data to a centralized server. There exists two types of federated learning:\\

\begin{itemize}
\item \textbf{Model-Centric Federated Learning} This is the most common of the two types.
In Model-Centric, the model is hosted in a cloud service, and its API is pre-configured (Layers, Weight, Etc ..). Each user downloads the model, enhances it and uploads a newer version. But this could happen over a long period of time, mostly weeks and months \cite{yang_federated_2019}.
Figure~\ref{fig:Model-Centric Federated Learning} describes its process. First, the model is sent to the edge device. Then, the metadata is sent to the global model, which resends an updated model.

A great example is Google's GBoard mobile app. It learns users typing preferences over time without sending any private data.

\begin{figure}[ht]
    \centering
    \includegraphics[width=0.35\textwidth]{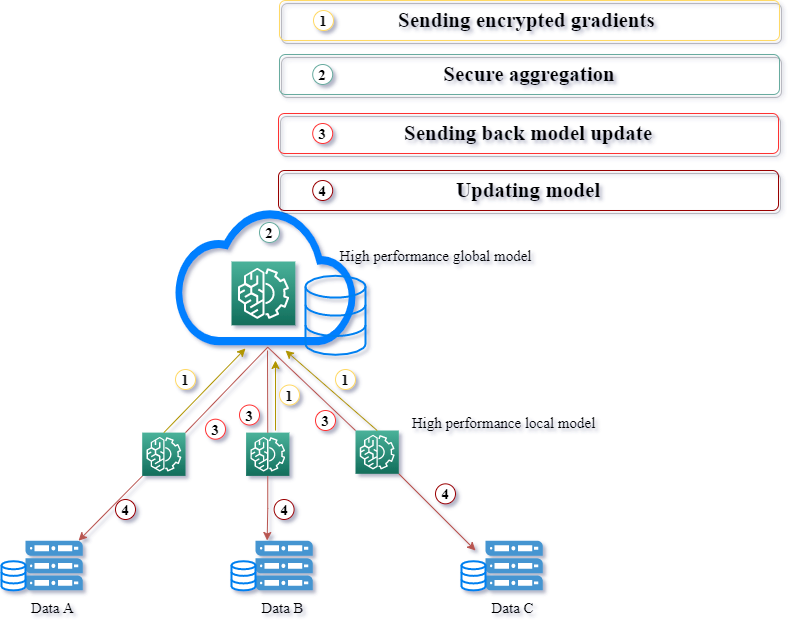}
    \caption{Model-Centric Federated Learning}
    \label{fig:Model-Centric Federated Learning}
\end{figure}

\item \textbf{Data-Centric Federated Learning} Less common than model-centric, but more suitable for experimentation science\cite{majcherczyk_flow-fl_2020}. It is hosted on a cloud server, not as a model but as a dataset whose API is also pre-configured (Schema, Attributes, etc ...). Users can use those datasets to train their model locally in a specially appointed practical way \cite{xie_multi-center_2020}. As described in figure~\ref{fig:Data-Centric Federated Learning}, the model owner sends the model to a cloud server asking for a specific type of data to train it. As soon the data is available and accepted by its owner, the model starts to train and eventually is sent back to its owner. 

For example, suppose an individual wants to train a model for a medical diagnosis. In that case, he needs to be part of a major hospital or constantly interact with a doctor. Otherwise, it may be highly troublesome or difficult to get a dataset that would be adequate for the model training. With data-centric, one could possess that data and add them to the cloud server for others to benefit from and train their model by submitting requests. 

\begin{figure}[H]
    \centering
    \includegraphics[width=0.4\textwidth]{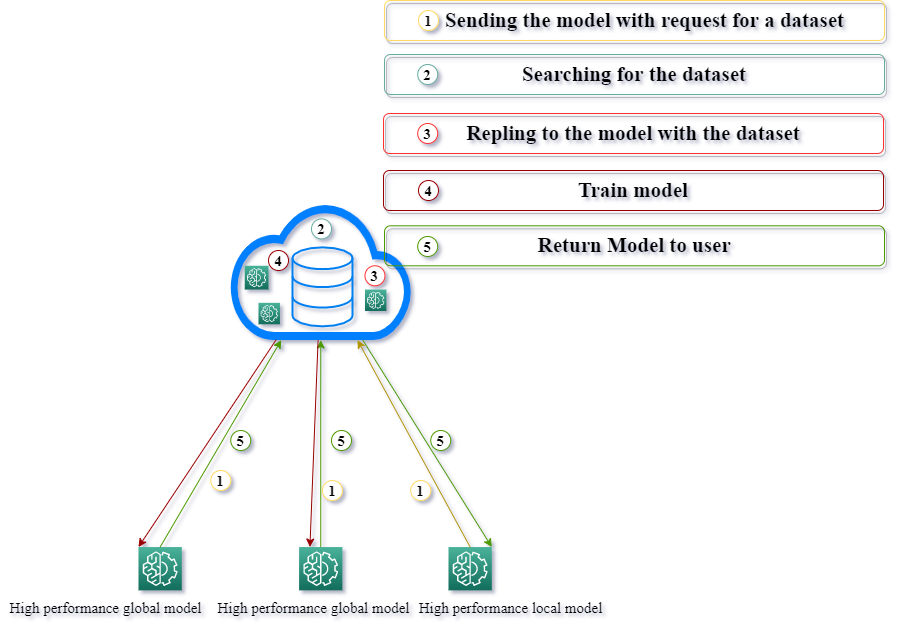}
    \caption{Data-Centric Federated Learning }
    \label{fig:Data-Centric Federated Learning}
\end{figure}
\end{itemize}

\section{Related Work}
This section discusses previous research papers related to the merging of blockchain technology, the machine learning process, and the integration of collaborative training. This idea is considered a new but promising one, considering the advantages it offers. Few works have been done in this area.\\
Generally, traditional Federated learning relies on a single central server to train a global model. If that server fails due to malicious behavior, all could be lost. Decentralization of the data and the training can provide a solution.

The research presented in the paper \cite{harris_decentralized_2019} focuses on improving machine learning model integrity and availability through decentralization. Their idea was to study the impact of blockchain technology over machine learning models, focusing on decentralization to mainly to provide sharing access to machine learning models. They used blockchain to establish a distributed build of large datasets accessible to all the network participants. Collaboration is a significant feature of their framework where they encourage contributors to improve and train a model constantly. Providing a continuously updated model, yet free, open and accessible is what this paper aims for.\\
The framework managed to create a distributed, collaborative environment where people could contribute effectively to a machine learning model using quality data.  This helped improve the model's accuracy, as it dealt with data manipulation thanks to a well designed incentive mechanism. The blockchain helped to create an open, secure, and with ownership datasets market where a participant could interact with to train models either on-chain or off-chain.\\
Unfortunately, it only manages to work with small models that can be efficiently updated with one sample. This approach of storing the entire dataset inside the blockchain is not very beneficial due the amount of data you can store. This is either because the amount is limited by the protocol, or because of the huge transaction fees you would have to pay. Therefore, the amount of data you store has to be held by every full node on the network. Everyone that downloads the blockchain is downloading your piece of data as well. Even keeping kilobytes can cost a fortune. 
When storing data on the blockchain, we often pay a base price for the transaction itself plus an amount per byte we want to keep. If smart contracts are involved, we also pay for the execution time of the smart contract \cite{hu_comprehensive_nodate}. With that being said, this option is not used to provide a collaborative environment to train complex models that need a large amount of data.\\

The work in the paper \cite{ma_when_2021} presents a secure and reliable federated learning framework. The idea is to ensure the model update process in a decentralized environment. Each client using the framework could train his machine learning model and mine blocks to publish aggregating results. The blockchain will track tasks related to collaborative learning, such as broadcasting learning models, publishing a task, or reviewing the aggregating learning results. Using decentralization accountability, they enable all miners to validate the uploaded model quality. This ensures misbehaving detection for low-quality contribution to the federated model. They start by integrating a local model into each training node, training and updating the model using a global model and its data. That leads to creating a training pool, conducting a decentralized aggregation process for the federated learning. Once the models are updated and collected, the clients calculate the global model update. Eventually, each block will record the local model update, data size, computation time, and aggregated parameters. Once registered, the model is published to the whole network. Once published and the hash value is found, miners need to verify the block contained aggregated results, either by comparing with the result found in the publishing block or by using a public dataset to justify the performance of the uploaded model. Reward allocation is available for the client to encourage them to mine and verify each block. Their study showed significant results for dealing with federated learning single server failure and avoiding unwanted distributed training behavior. However, the study shows that the data is available for the clients to train their model remotely and test its efficiency. This is not always the case.

Tao Wang \cite{wang_unified_2018} explored the establishment of a link between machine learning and blockchain technology to propose a unified framework aiming to deal with machine learning traceability and automation problems. It discussed the usage of multiple threads to translate core machine learning implementation instead of a single thread.
They used a three-layer architecture composed of a Server Layer, Streaming Layer, and a Smart Contract Layer. The server and streaming layers aim to achieve the seven machine learning steps (Initialization, Training, Validation, Testing, Evaluation, Serialization and Clean-up). Still, the first is using a static handle of data (data at rest) and the second a dynamic one (events on the fly). The Smart Contract layer is essential to deal with trustability and the automation of the process of training the model.\\ 
This work managed to achieve a stable framework to collect data using cloud technology and train machine learning models.\\ 
Nevertheless, the usage of smart contracts can cause runtime problems while running thousands of contracts on the blockchain, the response time can decrease tremendously.\\
\begin{figure*}[t]
    \centering
    \includegraphics[width=0.9\textwidth,]{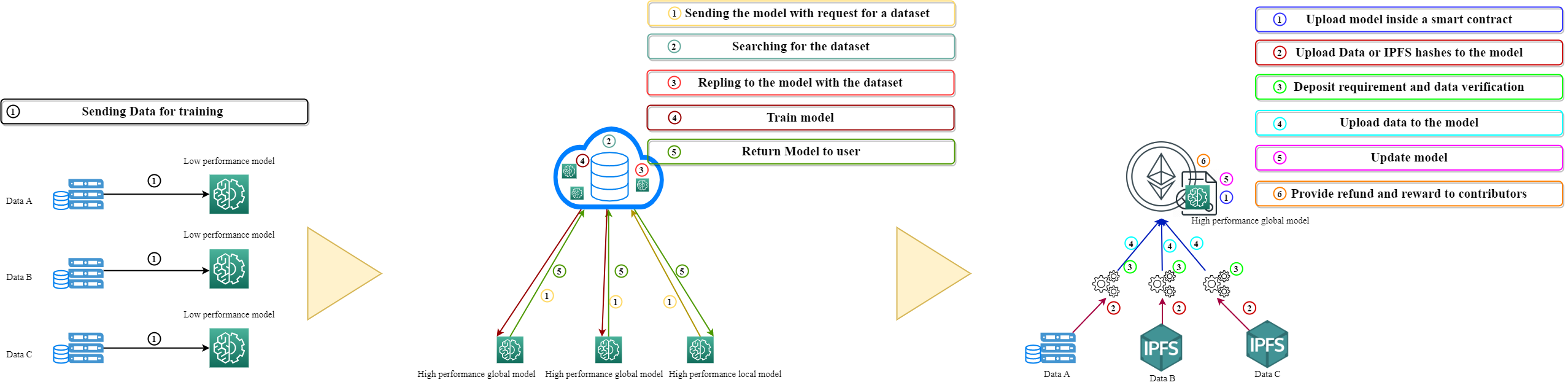}
    \caption{Our contribution overview : Blockchain integration for data-centric federated learning}
    \label{Blockchain integration for data-centric federated learning}
\end{figure*}
Each study has contributed to understanding the application of blockchain machine learning training, either in data availability or model training. As we mentioned previously, they provided the ability to conduct that contribution, such as automated model training, data availability, and verifying the quality and integrity of the data. 
But with each work comes limitations. We are still facing some related to the usage of blockchain technology. One of the main problems is data storage. The biggest problem of storing data on a blockchain is the amount of data you can store, either because the amount is restricted by the convention or given the immense exchange expenses you would need to pay.\\
One other major limitation is the need for high computation power to train a machine learning model in an efficient time matter. When we are using on­chain training with smart contracts, this usage can cause runtime problems while running thousands of contracts; the response time can decrease tremendously and even causes network failure.\\
In our study, we consider not only supervised models but also unsupervised ones with a similar incentive mechanism \cite{harris_decentralized_2019}. In this matter, we address the Etherieum \cite{vujicic_blockchain_2018} memory allocation problem. Our alternative solution is to use the InterPlanetary File System (IPFS) \cite{benet_ipfs_nodate} where we will only store on the blockchain the links of the file; this will help reduce the cost of storing inside Etherieum and allow people to upload as much data as they want. This will provide the ability to use diverse models since it will provide the necessary data regarding its training. And we offer the ability to train a model in an off-chain manner. This will allow the possibility of training models faster by avoiding network latency and then re-uploading them into the blockchain.

\section{ Proposed approach}
Distributed data is a critical element of our study since it provides a solution to data centralization. Data-centric Federated learning architecture was a reference approach in creating our architecture. While in federated learning, we distribute the model to the data in our study, we reverse it by spreading the data to the model. 
We explain in the figure below~\ref{Blockchain integration for data-centric federated learning}, the difference between two common types of learning, centralized learning and federated learning, and our framework contribution to model learning. 

\subsection{System Overview }\label{AA}

In classic centralized learning, we have data centralized in a server sent to a model for training. We spent decades using that technique until we were presented with federated learning. To be more specific, data-centric federated learning is the idea that a super or global oracle that is centralized in a server and contains multiple datasets added by contributors. Any model owner wanting to use this data could send their model and a request to use a specific dataset for its training. If accepted, the oracle searches for the requested dataset and starts the training of the model. As soon as the training is finished, the model is provided back to the model owner.\\
In the case of our framework, we have a model which is stored in a decentralized environment, being the public blockchain \cite{kim_blockchained_2020}. It is therefore accessible by all the network participants. Those same participants provide the data to the model in a collaborative approach. With every input of data or IPFS hash, we have data aggregation \cite{ramanan_baffle_2019}. But before any update to the model, the data needs to be verified to ensure its integrity. That function is provided using our incentive mechanism. The approved data is then sent to the model for training and the model is updated. For each beneficial deposit, a refund is made, and for each beneficial verification of the inputs, a reward is offered.
We are aiming more at data distribution rather than model distribution. Because that will help us create datasets for future usage. All of this is described in figure~\ref{Blockchain integration for data-centric federated learning}.
Our system aims to apply blockchain advantages in machine learning; by merging The Ethereum Blockchain and the InterPlanetary File System, we will be able to create a safe and public market of collaborative training and sharing of machine learning models. Smart contracts are the core of our system, where users will be able to upload machine learning models and IPFS hashes that relate to datasets stored in a decentralized manner. Other users could reach those attributes for improving them. One other feature is the ability to make IPFS hashes available to create a data-sharing system. Our research concerns mainly two areas of work: 
\begin{itemize}
	\item  Collaborative Federated Data centric
\end{itemize}
Where we present our strategy for reducing Blockchain storage costs so we could transform it into a data sharing and collection environment where people could have access to various data sets.
\begin{itemize}
 	\item  Collaborative Training 
\end{itemize}
Aiming to encourage users to add their machine learning models inside Ethereum blockchain via a smart contract, where contributors could interact with it to train, either by adding data or retrieving it and train the model in an off-chain manner before re-uploading.  

\subsection{Data Collection}\label{subsubsection}
\begin{figure}[H]
    \centering
    \includegraphics[width=0.55\textwidth]{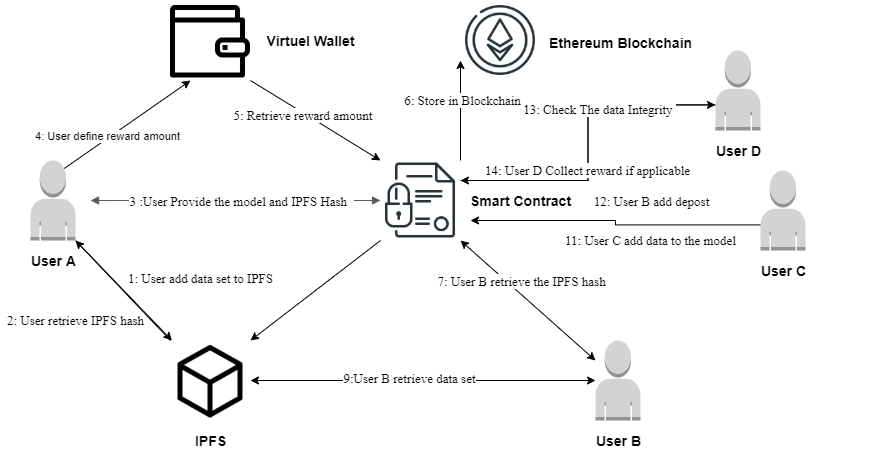}
    \caption{Data Collection Process}
    \label{fig:Collaborative Data Lake Process}
\end{figure}

Composed of the Data Handler and our Incentive Mechanism, those components aim to add and verify the integrity of the data added by the user to train the model.\\
The Data Handler collects and adds the data to the smart contract, which will be available to all participants of the network.\\
As for the incentive mechanism, it helps us ensure the quality of the data that is being distributed throw the blockchain by imposing a deposit when adding data and rewards for contributors when verifying it.\\ A smart contract is created and initialized to values of the incentive mechanism parameters. It then accepts actions from participants and triggers payments. Adding data requires validation from the incentive mechanism. The data will be stored in the smart contract either in the data handler or as an IPFS hash.
Figure~\ref{fig:Collaborative Data Lake Process} below describes the steps of the process of adding and collecting data to and from the smart contract in different scenarios. It consists of four users, a virtual wallet, the IPFS, the smart contract, and The Ethereum Blockchain.\\ 
\begin{itemize}
    \item \textbf{Step 1:} User A which is the initial user add his data file to IPFS
    \item \textbf{Step 2:} User A retrieves the hash of the file 
    \item \textbf{Step 3:} Start by providing the model, the data corresponding to the model, and the test data to the smart contract
    \item \textbf{Step 4:} User A defines the reward fees for contributors
    \item \textbf{Step 5:} The smart contract retrieves the amount and stores it
    \item \textbf{Step 6:} The smart contract is stored in the blockchain
    Those six steps represent "The model initialization steps"
    \item \textbf{Step 7:} User B retrieves the IPFS hash 
    \item \textbf{Step 8:} User B accesses IPFS platform and retrieves the file corresponding to the hash 
    \item \textbf{Step 9:} User C makes a deposit
    \item \textbf{Step 10:} User C adds data to the smart contract
    \item \textbf{Step 11:} User D makes a deposit, verifies and fixes malicious data 
    \item \textbf{Step 12:} User D retrieves a refund plus a reward fee  
\end{itemize}

\begin{figure}[ht]
    \centering
    \includegraphics[width=0.5\textwidth]{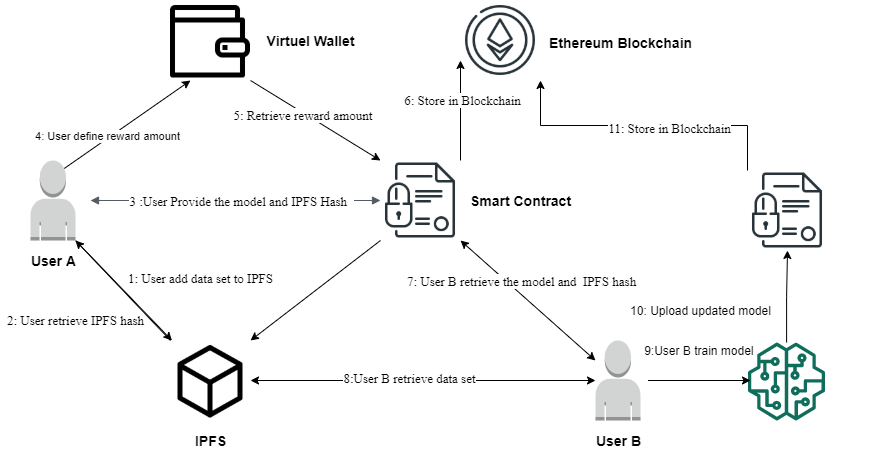}
    \caption{Model training process}
    \label{fig:Collaborative Training Process}
\end{figure}

\subsection{Model Training }
In order to automate the training process, users will generate a smart contract containing the machine learning model with all the specific details and its method of training. When data is received, it automatically starts the required action. When user A adds the model to the blockchain, it becomes public to other network users. When user B wants to contribute, there are two options; if the model isn't complex and could be updated with one sample, the user could manually add the input after acquiring a deposit. Alternatively, they needs to add the data set to IPFS, then add that hash to the contract. For the training to occur, attributes of the dataset are required to be similar to the initial ones used to train the model.\\
Figure~\ref{fig:Collaborative Training Process} above describes the steps of the process of training a model in different scenarios. It consists of three users, a virtual wallet, the IPFS, the smart contract, and the Ethereum Blockchain.\\

\textbf {The model initialization steps} 
\begin{itemize}
    \item \textbf{Step 7:} User B downloads the model and the IPFS hash 
    \item \textbf{Step 8:} User B accesses IPFS platform and retrieves the file corresponding to the hash
    \item \textbf{Step 9:} User B trains the model on his device
    \item \textbf{Step 10:} User B uploads the model in a new smart contract with references to the previous one  
    \item \textbf{Step 11:} User C makes a deposit and adds data to the model that triggers and update function   
    \item \textbf{Step 12:} The smart contract starts the training of the model using the new input and tests it using the test data   
\end{itemize}

\subsection{Implementation}
The whole system consists of three main components: back-end blockchain, IPFS storage, and front-end web UI. The client front-end will send a transaction to the blockchain to deploy a User contract for the current account. The user will be provided with an interface to upload the required files, machine learning model, and IPFS file. Previous research proved that blockchain could be suitable. We used Ethereum as our blockchain since it is public and supports smart contracts. On the other hand, IPFS is a distributed storage for our databases. The framework will store the hash of the IPFS file. 

\subsubsection{Data Hashing} The model owner breaks down the whole dataset into several data groups by using IPFS. It will create an IPFS object and generate a hash leading to the file address. This will allow us to benefit from large datasets without the need to store them inside the blockchain.\\
Moreover, the need to have a different hash to the test dataset is required in order to prevent it from being visible to the contributors. This help avoids overfitting problems. We used Sha-256 \cite{appel_verication_2015} as the hashing function for creating different hashed data groups for testing datasets. It is the same hashing algorithm used by IPFS.\\
\begin{figure}[h!]
    \centering
    \includegraphics[width=0.5\textwidth]{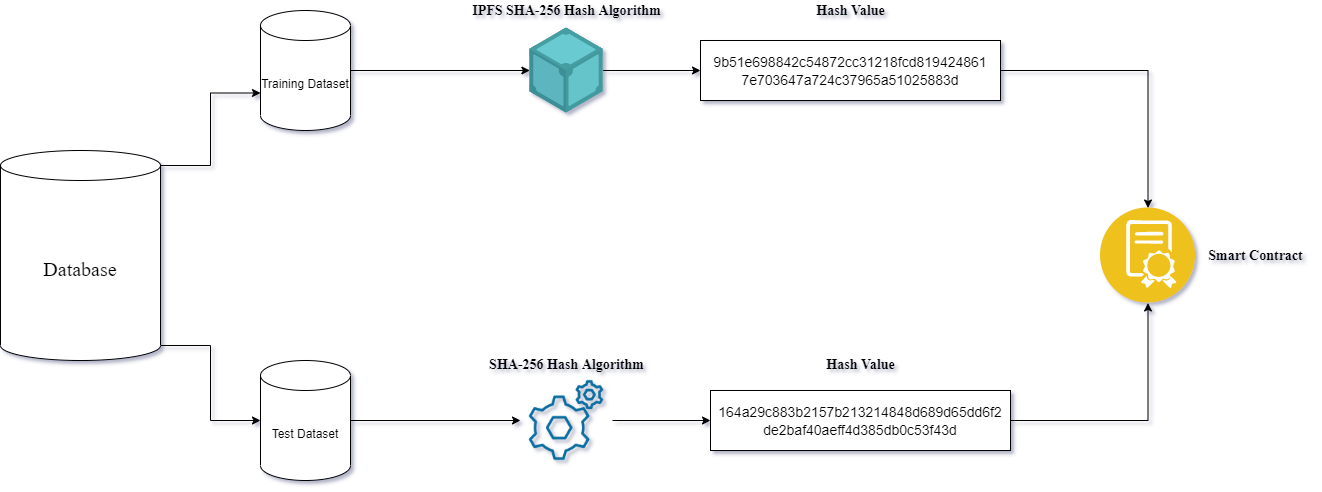}
    \caption{Hashing Process}
    \label{fig:Hashing Process}
\end{figure}

As mentioned in figure~\ref{fig:Hashing Process}, the user needs to split the the data into two datasets, training and testing. Moreover, he needs to run the training dataset throw the IPFS to generate the hash and eventually store it into the smart contract. As for the testing dataset, when added to the framework, a hash function will be generated using the SHA-256 algorithm to obtain a hash value. This value will also be stored in the smart contract.

\subsubsection{Adding Data}
\begin{figure}[h!]
    \centering
    \includegraphics[width=0.45\textwidth]{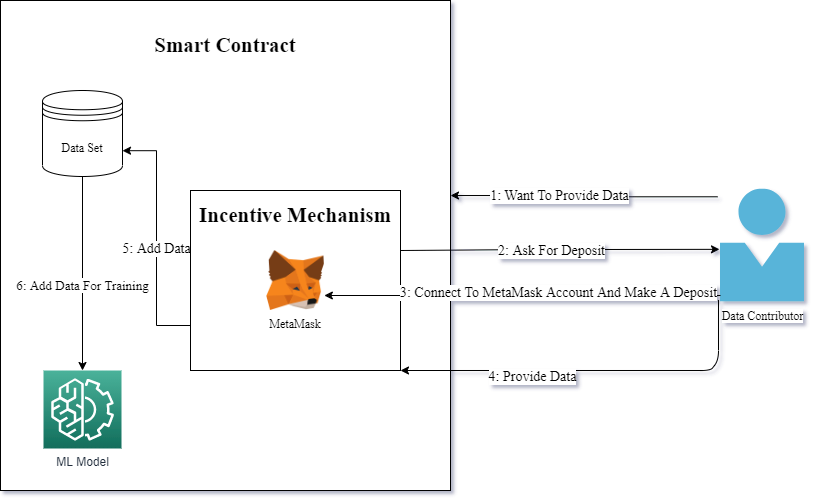}
    \caption{Incentive Mechanism: Data Adding }
    \label{fig:Incentive Mechanism: Data Adding}
\end{figure}

We call an add function that will allow the user to add data to the machine learning model either with simple input or as a hash of an IPFS file. This data will aim to train the model inside the blockchain. Those inputs will be later reviewed by contributors thanks to the incentive mechanism that will allow them to benefit from correcting and ensuring the data integrity. Other people could also download the data and use it to train their own machine models as described in figure~\ref{fig:Incentive Mechanism: Data Adding}.

\subsubsection{Data Integrity}
\begin{figure}[h]
    \centering
    \includegraphics[width=0.5\textwidth]{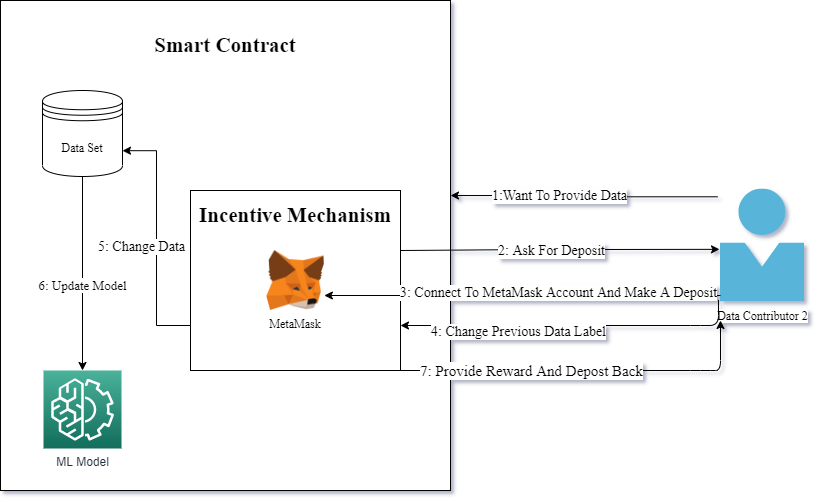}
    \caption{Incentive Mechanism: Data Verification}
    \label{fig:Data Integrity}
\end{figure}

To ensure the integrity of the data, we implement an incentive mechanism to encourage contributors and oblige each user to insert a deposit to add data to the model to make it costly for malicious users to disturb the model efficiency. It starts with the model owner providing a deposit, reward, and time-out function. It creates the fundamental aspect of our incentive mechanism. The deposit corresponds to the monetary amount that a user needs to make to add data to the smart contract to train the model on-chain. The reward represents the amount that a user will receive when checking the integrity of the data and restoring it. The time-out function defines the time between a return of the deposit and the validation of the owner.
Figure~\ref{fig:Data Integrity} describes the process of verification. Following a data input, a user has the opportunity to check it due to blockchain transparency.
An Interface shows all the contributors' transactions, meaning that all added data is visible for the owner and other contributors. If a data is wrong or does not fit the process of learning of the concerned model, he will be allowed to change it and update the model.\\ 

\subsubsection{Training Process}
The user will be provided with an interface where he creates his customized smart contract. The model owner adds his dataset inside IPFS using a framework interface to store it and then retrieves its corresponding hash. Next, they provide the machine learning model with some description of the model, the data used, and the hash of the IPFS inside the smart contract. Then, they define an amount related to the data verification reward. The smart contract will retrieve that sum from the user's virtual wallet and store it inside the Ethereum Blockchain. Network participants connect to the framework and access the smart contract to download the model and the IPFS hash to enhance the collaborative training. Using the hash, they obtain the dataset and start training the model off-chain. Finally, they add the updated model to a smart contract and store it in the Blockchain.
\section{Results}
Each user will have one separate individual smart contract. They will be able to upload a machine learning model and IPFS file containing the data used to train the model. Meanwhile, contributors will be able to access the smart contract to retrieve the model and the IPFS file. After training and improving the model, they could re-upload it inside the smart contract and the data they used for its training as an IPFS file. The usage of a public blockchain will provide more frequency of contributors willing to validate the data integrity. \\
The model will remain available inside the blockchain for future work with a view of its accuracy. This flow of interaction created an efficient environment for collaborative learning since it managed to make datasets available for all network users either with direct input inside the smart contract when we are dealing with simple inputs or with large ones when adding the file inside IPFS and sharing its hash in the blockchain. Their access will be at no cost. With the presence of the incentive mechanism, we managed to reduce the amount of ambiguous data that could be added to the model. The deposit function makes them costly and unworthy for malicious users. On the other hand, it encourages contributors since discovering those insufficient data benefits them with financial rewards. All of this improves the deployed model's accuracy and provides usable datasets to test and train models.
\subsubsection{Results' Validation}
\begin{figure}[h!]
\centering
\includegraphics[width=0.4\textwidth]{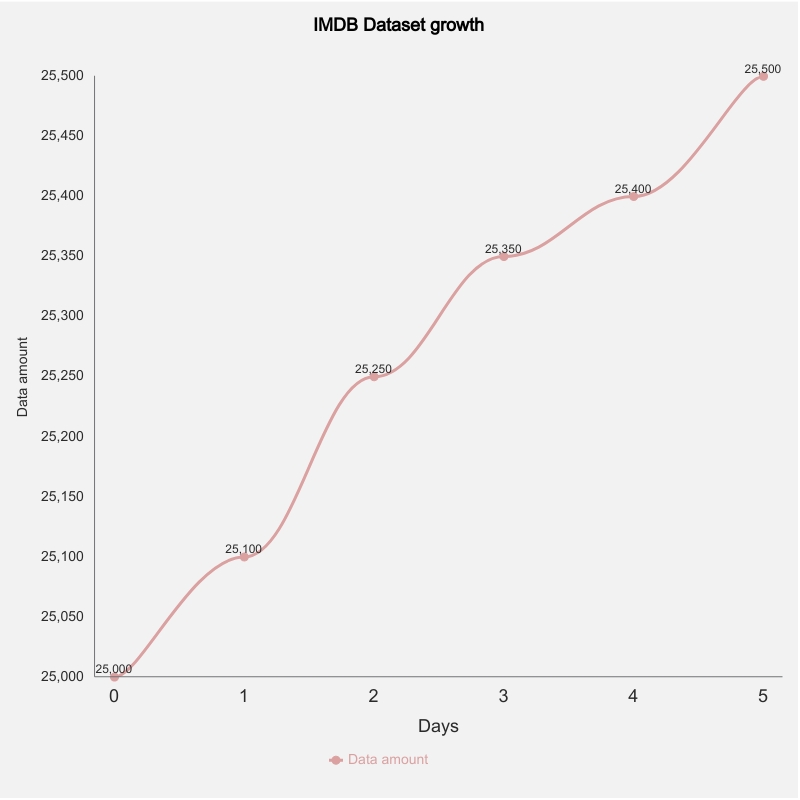}
\caption{IMDB dataset growth}
\label{fig:IMDB dataset growth}
\end{figure}

\begin{figure}[h]
\includegraphics[width=0.4\textwidth]{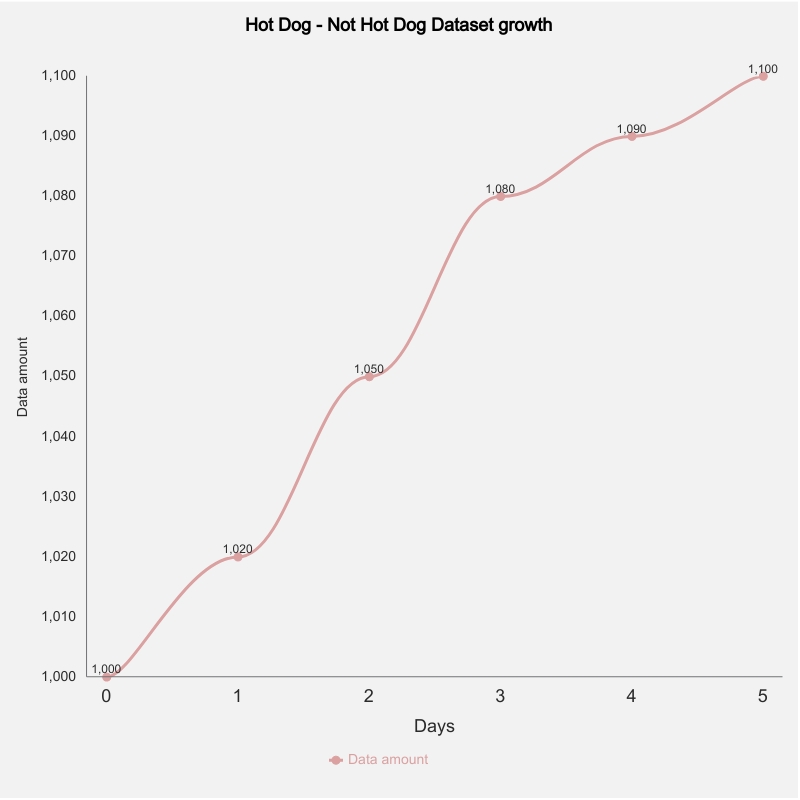}
\caption{Hotdog-not hotdog dataset growth}
\label{Dataset growth}
\end{figure}

\begin{figure}[h!]
    \centering
    \includegraphics[width=0.4\textwidth]{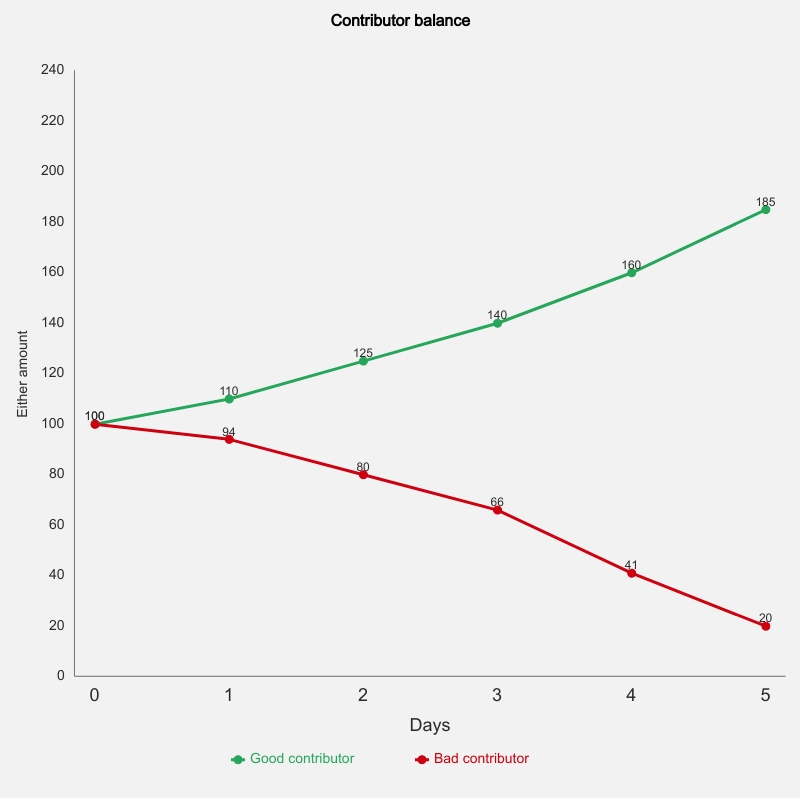}
    \caption{Agents Balance}
    \label{fig:agent balance}
\end{figure}

\begin{figure}[H]
    \centering
    \includegraphics[width=0.4\textwidth]{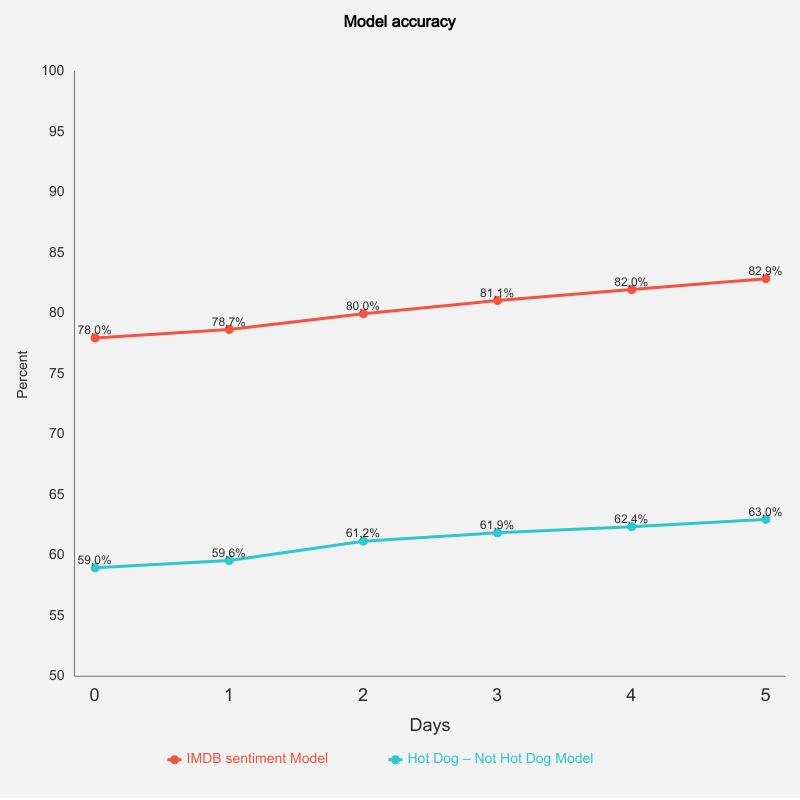}
    \caption{Accuracy evolution for the two models}
    \label{fig:model accurecy}
\end{figure}

Analyzing figure~\ref{fig:IMDB dataset growth}, we notice that we managed to obtain multiple inputs to our initial dataset. At the beginning of the experiment, the dataset contained 25,000 inputs of data. After using the framework, we observed that the number kept increasing until reaching 25,500 data inputs. An additional 500 data inputs were obtained in just six days with an average of 83 new inputs every day with only two contributors. That proves that our framework could add the data inputted from the contributors into the initial training dataset.
Same as the previous dataset, after analyzing figure~\ref{Dataset growth}, we notice that the amount of inputs in the training dataset had increased from 1000 in the initial state to 1100 input. That provides us with an additional 100 pictures in the dataset with their corresponding label.\\
In our framework, any user can participate and train a public model after providing a deposit in cases that deposit contains false data and leads to a lack of quality.\\
Our Incentive Mechanism (IM) was created to deal with malicious users that aim to disturb the model accuracy by providing incorrect inputs into to training dataset. By requiring a deposit with each input and a reward mechanism with each verification, we aim to stop those disturbing inputs and encourage other contributors to verify them. Our simulation has yield the following graph~\ref{fig:agent balance}.
 Figure~\ref{fig:agent balance} represents each contributor's ether balance, a good agent, and a malicious agent in their virtual wallet. We consider a malicious agent as a network participant who contributes to the model training with inadequate data. We can assume that there is a negative correlation between the two ratios. It is for the fact that after each deposit, the two agents add data to the contract, but while verifying the data, the good agent adjusts the incorrect data. Hence, he obtains a full refund plus a reward amount, while the other agent doesn't get his deposit. Each time the malicious agent kept adding data, his balance reduced until almost reaching zero.\\
As for the accuracy, starting with the IMDB sentiment model, it was initially trained on 20,000 of the 25,000 training data samples, with a model accuracy of 78\%.\\
We notice that despite the presence of the malicious agent, the presence of the IM, and a good agent, the model was able to train inside the smart contract improving from 78\% to 82.9\% level of accuracy. This proves that with this model, our framework was efficient in terms of improving the training of the model thanks to the constant contribution of data by the contributors.\\
As for the second model, The Hot Dog – Not Hot Dog Model, it was initially trained on the full dataset using the 1000 inputs. By constantly adding data and verifying its integrity, we managed to obtain the following result illustrated in figure~\ref{fig:model accurecy}.
 
We observed that those added data helped to improve the model accuracy from an initial 59\% to 63\%. Proving that our smart contract was capable of training the model and that the provided data were beneficial to the accuracy.\\

\subsection{Performance Evaluation}
Our framework is expected to deal with many users, so performance and scalability are major factors. However, at this moment, public blockchain throughput is very limited. We use the Ethereum blockchain, allowing around 15 transactions per second with 15-second blocktime. Also, \cite{daniel_ipfs_2022} states that IPFS instances possess limited bandwidth leading to low critical scalability of IPFS. Plus, a quantitative analysis \cite{shen_understanding_2019} toward IPFS I/O performance from a client perspective showed that operations are performing resolving and downloading result a bottleneck while reading a remote node by an IPFS client. So, scalability still necessitates future improvement for a public blockchain and IPFS to ensure trouble-free for extensive usage.

Nonetheless, when we tested our framework locally running in a single wired LAN, the execution was smooth and rapid, as described in table \ref{tab:table-one}. However, with a more significant number of nodes and a distributed wan approach, those results could tremendously alter time execution. In the current version, we use Ethereum as the basic framework; as we run the experimentation locally, we didn't consider the influence of malicious node
behaviors on data quality. As for deploying the framework on the public Ethereum network, we recommend including machine learning-based detection of malicious Ethereum entities as described by Poursafaei et al \cite{poursafaei_detecting_2020}. 

\begin{table}[]
\begin{threeparttable}
\begin{tabular}{|l|c|}
\hline
\multicolumn{1}{|c|}{Operation}                                               & Time of Execution in seconds \\ \hline
IPFS Node Creation                                                            & 2 - 3                        \\ \hline
Smart Contract Creation                                                       & 15 - 20                      \\ \hline
IPFS Hash Upload                                                              & 2 - 3                        \\ \hline
\begin{tabular}[c]{@{}l@{}}Consensus \&\\  Ledger Update\end{tabular}          & 2 - 3                        \\ \hline
Data Adding                                                                   & 1 - 2                        \\ \hline
\begin{tabular}[c]{@{}l@{}}Deposit Payment \& \\ Update\end{tabular}           & 2 - 3                        \\ \hline
\begin{tabular}[c]{@{}l@{}}Model Training \\ With Single Input\end{tabular}   & 1 - 2                        \\ \hline
\begin{tabular}[c]{@{}l@{}}Model Training \\ With Multiple Input\end{tabular} & 60 - 80*          \\ \hline
Smart Contract Update                                                         & 2 - 3                        \\ \hline
Reward Payment                                                                & 1 - 2                        \\ \hline
Model Download                                                                & 1 - 2                        \\ \hline
Dataset Download                                                              & 10 - 15*                  \\ \hline
                                                                              
\end{tabular}
\begin{tablenotes}[para]
        \footnotesize
        \item[*] Can vary tremendously depending the size
    \end{tablenotes}
\end{threeparttable}
\caption{\label{tab:table-one}Framework process time consumption.}
\end{table}
\section{Discussion}
The study discussed here relates to creating a decentralized environment for collaborative learning of machine learning models. This research seeks to provide public access to datasets, share them, and automate the process of training models. The results acquired prove that leveraging Blockchain and IPFS offer a solid case to achieving our objectives.\\
The public blockchain makes such intuitively possible since it offers transparency access to stored data sets by storing their IPFS hashes and smart contracts. We were able to deploy our machine model, thus automating the training process. In addition, IPFS allows the shared files to be decentralized, and it provides a hash specific to that file to access it. Moreover, the results showed that it is non-beneficial and costly to add ambiguous data to the smart contract with the incentive mechanism since it keeps asking for a deposit with each iteration. Previous research demonstrated that blockchain and smart contracts are favorable for training non-complex machine learning models without a higher complex model since storing data in a public blockchain is very costly.
However, with the presence of IPFS, we could store those data without actually storing them and that by only storing the hash of the file, hence the training of more complex models.\\  
Our framework manages to create a dissent-sharing environment of machine learning resources and an automated training ground for all models.  This is considered one of the most considerable system contributions since we managed to train highly complex machine learning models while avoiding Ethereum storage's high-cost. 

\subsubsection{Limitations:}
Some limitations constrained the methodological choices; with the usage of IPFS, despite the multiple datasets, the incentive mechanism recognizes the file as only one data input, which will cost one date deposit and one data reward for the contributors. So despite the work put in the reward will not be significant.\\
Additionally, we didn't manage to implement the DanKu protocol \cite{kurtulmus_trustless_2018} due to dependency issues, which cost our system not to evaluate all the updated models inside the same smart contract, leading to creating a new smart contract with each update of the model. Furthermore, it is beyond the scope of this study to acknowledge the required computation power for on-chain training of the model, which could lead to low training efficiency.

\section{Conclusion}

Our research contributed to creating a decentralized public framework to enhance collaborative learning of the machine learning model on the Ethereum blockchain. The purpose is to provide specialists seeking contribution a network where they could find the right resources to accomplish their work. Data availability is crucial to the improvement of machine learning; hence, they aren't accessible; thus, this framework could be beneficial for such situations. Training also is a crucial part of our objective. Having the opportunity to train a model publicly is helpful to obtain a higher efficiency and accuracy rate. With their anonymous and distributed nature, smart contracts create a marketplace of model and data sharing without the need for intermediaries. With the association of an Incentive mechanism, quality insurance could be met.\\
Future studies should consider improving audit mechanism to optimize the data verification process and eradicate any potential corruption in the network. Also, we will investigate sharing computation power inside the blockchain to enhance the training of highly complex models such as deep learning since GPU needs are becoming essential in training. Blockchain could provide a sharing computation power network where users could share their computation resources inside the blockchain via a smart contract to train on-chain models.

\bibliography{aaai22}

\end{document}